\documentclass[letterpaper, 10 pt, conference]{ieeeconf}  

\IEEEoverridecommandlockouts                              

\usepackage{graphicx}
\usepackage{bm}
\usepackage{cite}
\usepackage{float}

\title{\LARGE \bf
Contact Localization through Spatially Overlapping \\Piezoresistive Signals
}

\author{Pedro Piacenza$^{1,4}$, Yuchen Xiao$^{1,4}$, Steve Park$^{2,3}$, Ioannis Kymissis$^{2}$ and Matei Ciocarlie$^{1}$
\thanks{$^{1}$Department of Mechanical Engineering, Columbia University, New York, NY 10027, USA.}%
\thanks{\hspace{-3mm}{\tt\small \{pp2511,yx2281,matei.ciocarlie\}@columbia.edu}}%
\thanks{$^{2}$Department of Electrical Engineering, Columbia University, New York, NY 10027, USA.{\tt\small johnkym@ee.columbia.edu},}%
\thanks{\hspace{-3mm}{\tt\small spark928@alumni.stanford.edu}}%
\thanks{$^{3}$Department of Materials Science and Engineering, Korea Advanced Institute of Science and Technology, Daejeon, Korea}%
\thanks{$^{4}$These authors contributed equally to this work.}%
}

\begin{document}

\maketitle
\thispagestyle{empty}
\pagestyle{empty}

\begin{abstract}
Achieving high spatial resolution in contact sensing for robotic
manipulation often comes at the price of increased complexity in
fabrication and integration. One traditional approach is to fabricate
a large number of taxels, each delivering an individual, isolated
response to a stimulus. In contrast, we propose a method where the
sensor simply consists of a continuous volume of piezoresistive
elastomer with a number of electrodes embedded inside. We measure
piezoresistive effects between all pairs of electrodes in the set, and
count on this rich signal set containing the information needed to
pinpoint contact location with high accuracy using regression
algorithms. In our validation experiments, we demonstrate
submillimeter median accuracy in locating contact on a 10mm by 16mm
sensor using only four electrodes (creating six unique pairs). In
addition to extracting more information from fewer wires, this
approach lends itself to simple fabrication methods and makes no
assumptions about the underlying geometry, simplifying future
integration on robot fingers.

\end{abstract}

\section{INTRODUCTION}

Tactile sensing modalities for robotic manipulation have made great
strides over the past years. Recent surveys list numerous such
methods: piezoresistance, piezocapacitance, piezoelectricity, optics,
ultrasonics, etc. Still, these advances in sensing modalities are only
slowly translating to improved abilities for complete robotic
manipulators. A possible reason is that the gap between an individual
taxel and a touch-sensitive hand often proves difficult to cross. As a
recent survey concludes: ``Instead of inventing `yet
another touch sensor,' one should aim for the tactile sensing system.
While new tactile sensing arrays are designed to be flexible,
conformable, and stretchable, very few mention system constraints like
[...]  embedded electronics, distributed computing, networking,
wiring, power consumption, robustness, manufacturability, and
maintainability.''~\cite{DAHIYA10}

In this paper, we focus on the problem of using a touch sensing
modality to achieve high-resolution sensing over relatively large
areas. Traditionally, this can be achieved using arrays of individual
taxels. However, this approach requires at least one wire per taxel,
or likely two; thus, an $m$-by-$n$ taxel array requires $2mn$
wires. At best, matrix addressing can achieve a similar result with
$m+n$ wires, but imposes regular geometry on the sensor. Individual
taxels must also be isolated from each other, further increasing
manufacturing complexity. We are motivated by finding a method to
achieve high resolution tactile sensing that circumvents these
difficulties and is intrinsically amenable to addressing other
system-level aspects from the list above, such as easy manufacturing
and coverage of unspecified geometry. In other words, \textit{we are
  looking for simple ways to achieve high-resolution tactile sensing
  with few wires}.

\begin{figure}[t]
\centering
\includegraphics[width=0.9\linewidth]{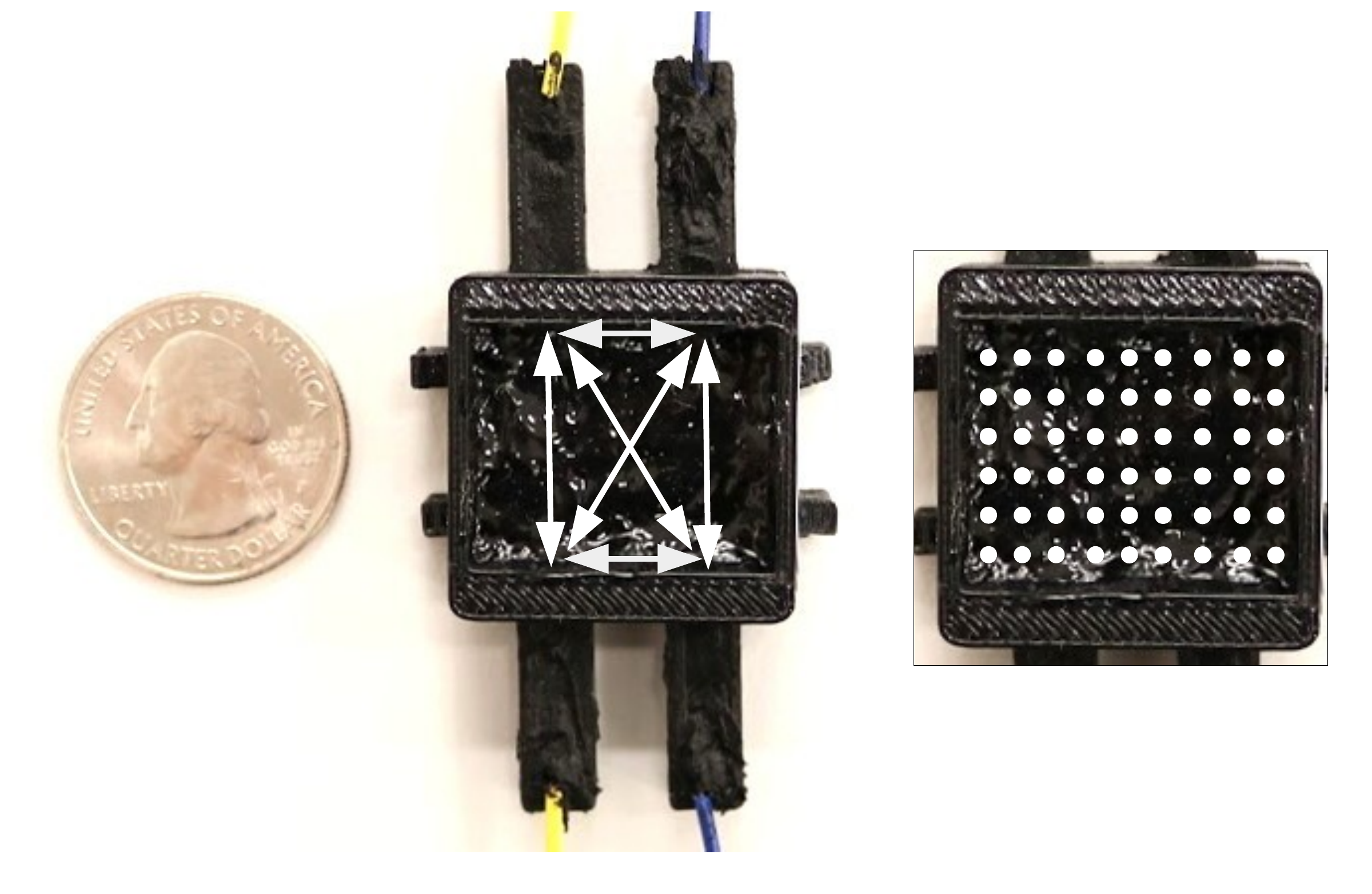}
\caption{Proof of concept design, with a rectangular volume of
  piezoresistive elastomer and four connected electrodes. We measure
  resistance change due to indentation between all electrode pairs
  (illustrated by arrows) and use grids of known measurements
  (illustrated on the right) to learn a mapping from these signals to
  indentation location.}
\label{fig:overall}
\end{figure}

Our approach, illustrated in Figure~\ref{fig:overall}, is to build a
single, continuous volume of a piezoresistive polymer with multiple
electrodes embedded in this volume. For an indentation anywhere on the
surface of this volume, we measure resistance change in response to
strain between all the pairs of electrodes in the area. This provides
us with a number of signals that is quadratic in the number of
wires. In addition to providing many signals with few wires, this
approach has other manufacturability-related advantages: the sensor is
a continuous volume that can be poured into a mold of arbitrary
geometry. No insulation between various components is necessary.

However, none of these advantages are meaningful if the information
carried by the signals from the many pairs of electrodes is not
sufficiently discriminative. The price paid for our approach is that
the relationship between each signal and the variables of interest is
difficult to determine analytically. We thus use purely data-driven
techniques and learn this mapping directly from indentation tests. The
large number of electrode pairs provides us with a many-to-few mapping
to variables of interest, a mapping that can be effectively learned,
as we show in this paper.

Overall, the main contributions of this paper are as follows:
\begin{itemize}
\item We introduce a new method for localizing contacts on a touch
  sensor by measuring resistance changes between multiple, spatially
  overlapping electrode pairs. This method lends itself to simple
  fabrication methods well suited for covering non-flat geometry.
 \item We demonstrate submillimeter median accuracy in determining
   contact position on a sensor with a 160mm$^2$ effective area. This
   is achieved using only four wires that connect to the sensor,
   creating six electrode pairs, and without relying on a flat rigid
   substrate or circuit board.
\end{itemize}

\section{Related Work}

Numerous types of transduction principles have been explored during
the last two decades when building tactile sensors: resistive,
capacitive, optical, ultrasonic, magnetism-based, piezoelectric,
tunnel effect sensors, etc. We refer the reader to a number of
comprehensive reviews~\cite{DAHIYA10,HAMMOCK13} for an overview of
these methods. Our goal however is not to explore a new sensing
modality; rather, we are looking to build on top of one such method,
in order to improve accuracy without sacrificing manufacturability.

Our basic building block is an elastomer with dispersed conductive
fillers applied to achieve piezoresistive characteristics. Numerous
examples of using this method exist in the
literature~\cite{ALEX14,DUSEK14,LIPOMI11,KIM15}, with carbon black and
carbon nanotubes as the most commonly employed fillers. Multi-layered
designs or additional microstructures can further improve
performance~\cite{MANNSFELD10,PARK14,WU15}. Recently, embedding
microchannels of conductive fluids inside an elastic
volume~\cite{PARK12,VOGT13} has been shown to be an effective
alternative to making the entire volume conductive, especially if
large strains are desirable. Here however we opt for the simplicity of
single volume isotropic materials which can be directly molded into
the desired shape.

Regardless of the base transduction principle, attempts to
increase spatial resolution have often resulted in the arrangement of
multiple discrete sensors into a matrix to cover a given target
surface. Some of these arrays can develop very high spatial
resolution~\cite{KANE00, TAKAO06, SUSUKI90}. However, a drawback of
this approach is the difficulty involved in manufacturing these arrays
onto a flexible substrate than can conform to complex
surfaces. Possible technologies to overcome this problem include
organic FETs/thin film transistors realized on elastomeric substrates
and other related techniques~\cite{SOMEYA04, SHIMOJO04, KIM08}. Still,
wiring and manufacturing complexity, along with other system-level
issues such as addressing and signal processing of multiple sensor
elements, remain important roadblocks on the way to building complete
sensing systems.

Some of these problems have been recognized and tackled before using
super-resolution techniques to reduce the amount of sensing units
needed while still achieving high resolution tactile sensing. Van den
Heever et al. ~\cite{HEEVER09} used a similar algorithm to
super-resolution imaging, combining several measurements of a 5 by 5
force sensitive resistors array into an overall higher resolution
measurement. Lepora and Ward-Cherrier\cite{LEPORA151} and Lepora et
al.\cite{LEPORA152} used an array of sensors covered in soft silicon
foam, such that the sensitive receptive fields of each individual
sensor overlap. Using a Bayesian perception method, they obtained a
35-fold improvement of localization acuity (0.12mm) over a sensor
resolution of 4mm.

A very closely related technique to the one presented in this paper is
the electrical impedance tomography (EIT)~\cite{HOLDER04}. EIT is used
to estimate the internal conductivity of an electrically conductive
body by virtue of measurements taken with electrodes placed on the
boundary of said body. While originally used for medical applications,
EIT techniques have been applied successfully for manufacturing
artificial sensitive skin for robotics
~\cite{NAGAKUBO07,KATO07,TAWIL11,ALIREZAEI09}. However, spatial
resolution of EIT-based skins has lagged behind other tactile sensing
technologies.

Our approach is to maintain the manufacturability and simplicity of
single-volume piezoresistive materials, while attempting to harvest a
large number of signals from many pairs of electrodes embedded in the
volume. We do not aim for analytical characterization of these
signals; rather, we aim to directly learn the mapping between this
data and our variables of interest. We note that machine learning for
manipulation based on tactile data is not new. Ponce Wong et
al.~\cite{PONCE14} learned to discriminate between different types of
geometric features based on the signals provided by a previously
developed~\cite{WETTELS08} multimodal touch sensor. Current work by
Wan et al.~\cite{WAN16} relates tactile signal variability and
predictability to grasp stability using recently developed MEMS-based
sensors~\cite{TENZER14}.

With traditional tactile arrays, Dang and Allen~\cite{ALLEN13}
successfully used an SVM classifier to distinguish stable from
unstable grasps in the context of robotic manipulation using a Barrett
Hand which provides tactile feedback through four arrays of 24
taxels. Bekiroglu et al.~\cite{BEKIROGLU11} also studied how grasp
stability can be assessed based on tactile sensory data using
machine-learning techniques like AdaBoost, SVM and HMM with similar
success. In similar fashion, both Saal et al. \cite{SAAL10} and Tanaka
et al.~\cite{TANAKA14} used probabilistic models on tactile data to
estimate object dynamics and perform object recognition
respectively. However, most of this work is based on tactile arrays
built on rigid substrates and thus unable to provide full coverage of
complex geometry. In contrast, we apply our methods to the design of
the sensor itself, and believe that developing the sensor
simultaneously with the learning techniques that make use of the data
can bring us closer to achieving complete tactile systems.

\section{Sensor Design and Construction}

\begin{figure}[t]
\centering
\includegraphics[width=0.7\linewidth]{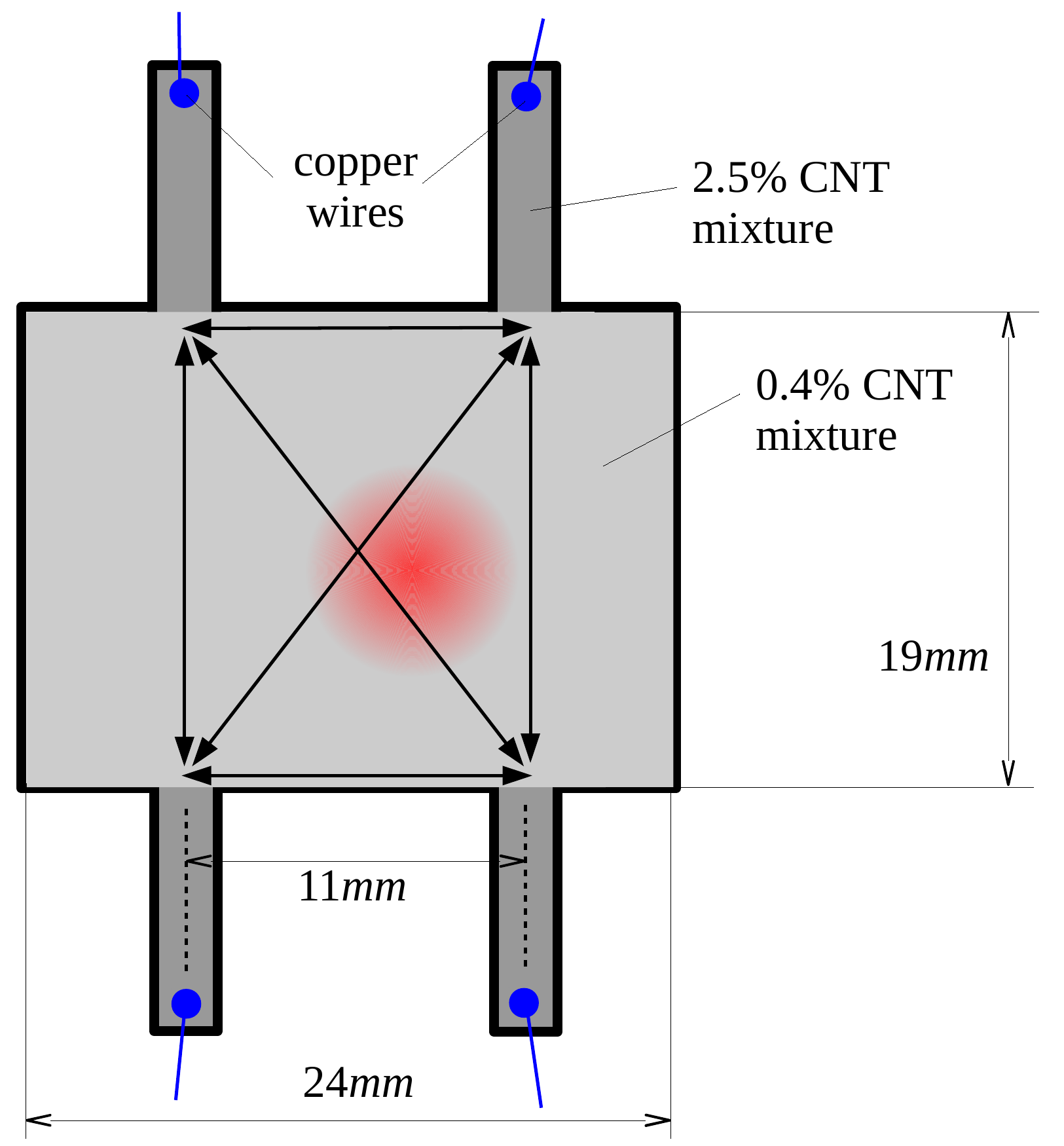}
\caption{Design of the proof of concept sensor used for the
  experiments in this paper. The rectangular center is filled with
  piezoresistive PDMS/CNT mixture. Side channels are filled with a
  conductive mixture with higher CNT ratio in order to mechanically
  isolate copper wire contacts from indentations. For a given
  indentation (illustrated by red circular pattern) we measure the
  change in resistance between all six electrode pairs (illustrated by
  arrows on the surface of the volume).}
\label{fig:sensor}
\end{figure}

Our overall approach to achieving high spatial resolution starts with
a continuous volume of piezoresistive material, with a number of
embedded electrodes. Molding a silicone elastomer into the
desired shape allows us to embed electrodes while the mixture is
viscous, and keeps open the possibility of covering complex, non-flat
surfaces in the future. To achieve piezoresistance for our silicone,
we use the well established method of dispersing a conductive filler,
as detailed below. We then describe the switching circuit developed so
that we can sample the change in resistance between any pair of
electrodes at high rates.

\subsection{Building a Piezoresistive Material}

We disperse multiwall carbon nanotubes (MWCNT, purity: 85\%, Nanolab
Inc.) into polydimethylsiloxane (PDMS, Sylgard 184, Dow Corning), a
two-part silicone elastomer. The key aspect of this process for
achieving piezoresistance is choosing the appropriate ratio of
conductive filler to elastomer. According to the commonly used
percolation theory, the conductivity of the composite w.r.t. filler
ratio displays an inflection near a point referred to as the
percolation threshold. A composite with that ratio will also display
the most pronounced piezoresistive effect~\cite{HU11}. In order to
find the percolation threshold of our materials, we built and tested a
series of samples with the concentrations of MWCNTs from 0.2wt.\% to
5wt.\%. We found that the most pronounced change in conductivity
occurred around the threshold of 0.4wt.\% filler, which we used in all
subsequent experiments.

In order to achieve uniform distribution of carbon nanotubes within
PDMS, we use a chloroform as a common solvent, an approach referred to
as the solution casting method~\cite{LIU12}. First, we add chloroform
and MWCNT into a beaker and sonicate with a horn-type ultrasonicator
in a pulse mode with 50\% amplitude for 30 min to evenly disperse
MWCNTs into chloroform. After that, we pour PDMS into the beaker (at
chloroform:PDMS weight ratio of 6:1 or more to reduce the viscosity
of the whole mixture), stir the mixture for 5 min to diffuse the PDMS
into the solvent, and then sonicate again for 30 min to disperse the
MWCNTs into PDMS. We then heat the mixture at 80$^{\circ}$ C for 24 hours
to evaporate the chloroform. After adding the curing agent, the
mixture is ready to be poured into the mold; for the experiments
presented here we empirically selected a curing agent to PDMS ratio of
1:20. Finally, the sample is finished after curing in an oven at
80$^{\circ}$ C for 4 hours.

\begin{figure}[b]
\centering
\includegraphics[width=\linewidth]{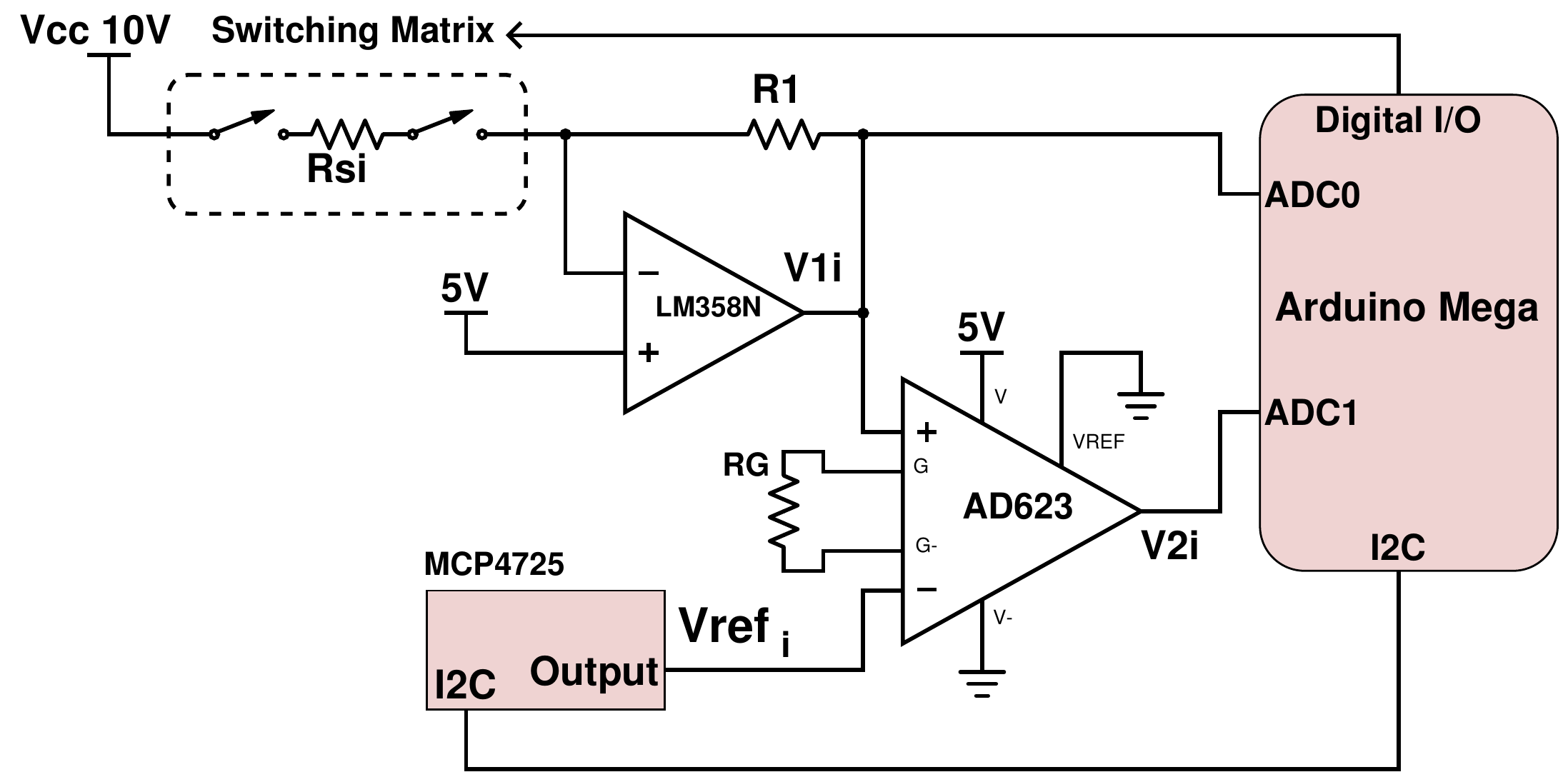}
\caption{Measuring circuit. For every terminal pair with a resistance
  $R_{si}$, we take a measurement of $V_{1i}$ with the sample at
  rest. This measured voltage is then reproduced on ${V_{ref}}_i$ by
  means of a digital to analogue converter such that we only amplify
  the change on $V_{1i}$ when the sample is indented.}
\label{fig:circuit}
\end{figure}

Our goal for this design is to measure resistance through a volume
between multiple pairs of terminals. However, in order to isolate
piezoresistive effects from mechanical changes at the contacts due to
indentation, we mechanically separated the wire contacts from the
piezoresistive sample placed under indentation tests. We extended a
number of 30mm side channels from the sample, each filled with a
CNT-filled PDMS mixture with a higher concentration of 2.5wt.\% We
then embedded copper wires directly into the mixture at the end of
these channels (Figure~\ref{fig:sensor}). The mixture with the ratio
of 2.5wt.\% has no piezoresistive characteristics and its conductivity
is close to that of the copper wires; thus, the mixture with the ratio
of 0.4wt.\% located at the center of the mold dominates the overall
conductivity.

\subsection{Sampling Circuit}

\begin{figure}[t]
\centering
\includegraphics[width=0.9\linewidth]{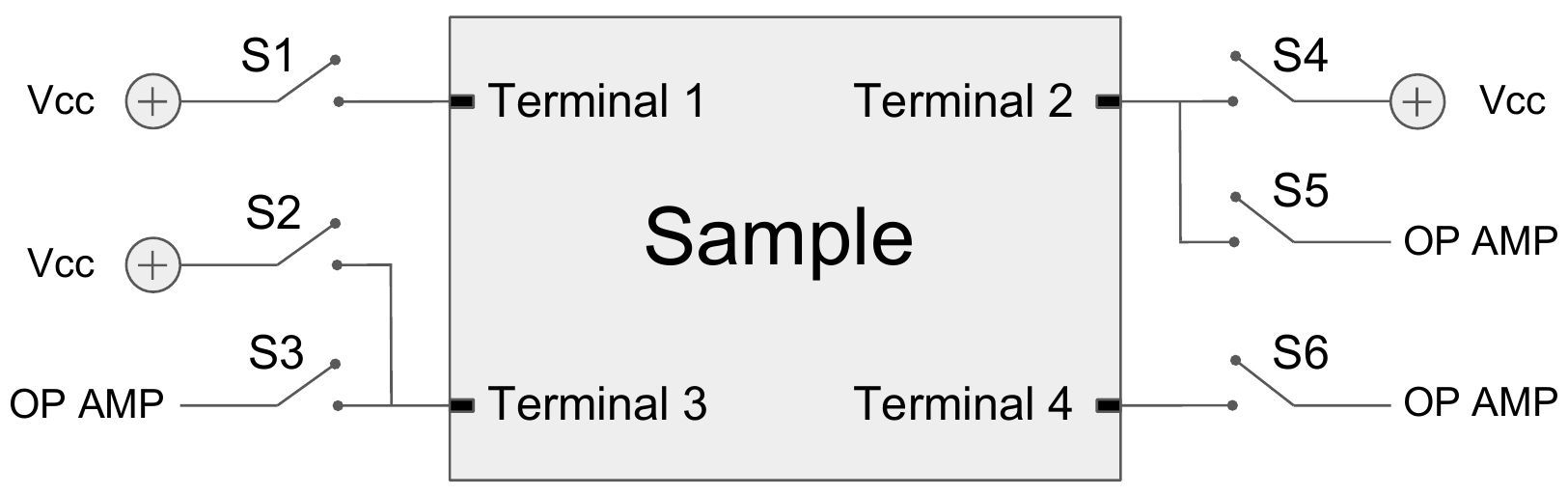}
\caption{Switching Matrix structure. All six switches are controlled
  with digital signals from our microcontroller. This configuration
  allows us to measure the resistance change across any pair of
  terminals. The switches must be closed such that we always connect
  one terminal to Vcc and the remaining one to the inverting input of
  the operational amplifier.}
\label{fig:switching}
\end{figure}

The main objective for our sampling circuit is to measure the change
in resistance between all pairs of electrodes/terminals that occurs as
a result of some strain being applied to our sample material. We found
that this change can be around 5\% of the nominal value at
rest. Furthermore, every pair of terminals will have different
resistance values at rest. It is also important to be able to sample
these relatively small changes in resistance at a high enough rate
such that a set of all measurements across terminals can be
representative of the instantaneous strain applied.

Consider $R_{si}$ to be the resistance across the $i$-th terminal pair
that we are interested in measuring, $i \in \{1,2,3,4,5,6\}$. As shown
in Figure~\ref{fig:circuit}, we use a first stage with a simple
operational amplifier in inverting configuration in a way that we
guarantee an output $V_{1i}$ between 0 and 5 volts such that it can be
directly measured by our microcontroller analogue to digital converter
module (ADC). The output of this stage, $V_{1i}$, is given by equation
\ref{eq:V1}

\begin{equation} \label{eq:V1}
V_{1i} = 5\texttt{V} - 5\texttt{V}\left(\frac{R_1}{R_{si}}\right)
\end{equation}

Since the change in the resistance $R_{si}$ can be very small, it
stands to reason that the change in our output voltage $V_{1i}$ from
this first stage will also be small. It is worth noting that the value
of $R_1$ has to be smaller than any of our values $R_{si}$, and the
sensitivity of $V_{1i}$ with respect to $R_{si}$ changing increases as
the value of $R_1$ is closer to those of $R_{si}$.

Because we are not interested in the absolute value of $V_{1i}$ but
only in its change over time when strain is applied, we take a baseline
measurement of $V_{1i}$ when the sample is undisturbed. These baseline
measurements are then used as the values of ${V_{ref}}_i$ that we hold
on the negative input of an instrumentation amplifier for the second
stage of the circuit. This allows us to remember the undisturbed value
of $V_{1i}$, compare it with the current one, and amplify that
difference. The voltage ${V_{ref}}_i$ is provided in our circuit by a
digital to analogue converter (DAC, MCP4725).

Our switching matrix is shown in Figure ~\ref{fig:switching}, where we
have six switches total, allowing us to measure across any combination
of terminals, T1 through T4. Each one of these terminal pairs has a
value of $R_{si}$, where $i \in \{1,2,3,4,5,6\}$. The overall circuit
delivers the set of all six $V_{2i}$ values every 25 milliseconds,
resulting in a 40Hz sampling frequency. This value is deliberately
conservative, since the only bottleneck on how fast we can switch our
matrix is down to the speed of the ADC module.

\section{Data Collection Protocol}

\begin{figure}[t]
\centering
\includegraphics[width=0.95\linewidth]{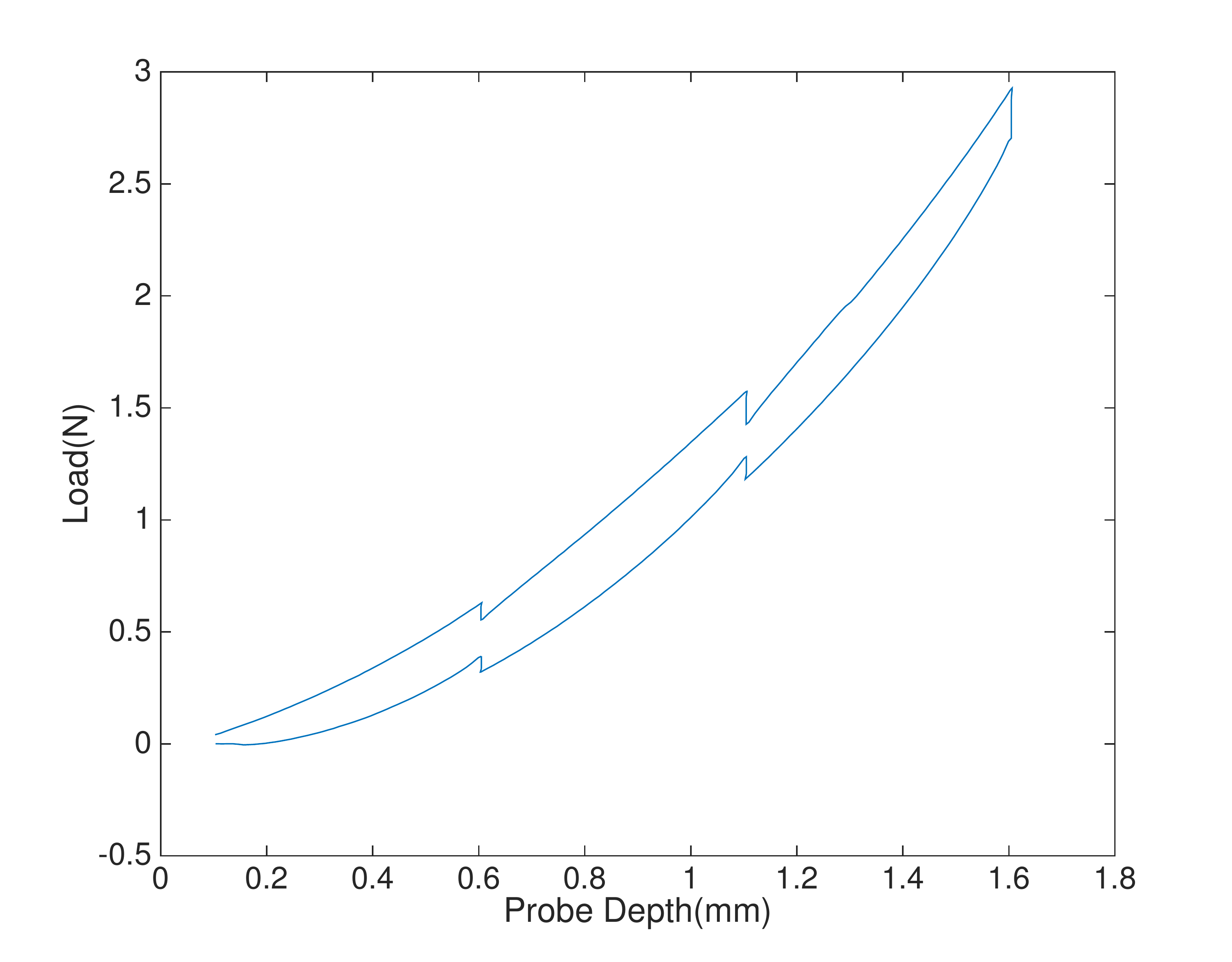}
\caption{Load vs. indentation depth for our CNT-filled PDMS samples,
  in loading and unloading regimes. Measurements were taken by
  advancing or retracting the probe in steps of 0.5mm separated by 30s
  pauses.}
\label{fig:instron}
\end{figure}

To collect training and testing data, we indent the sample at a series
of known locations and to a known depth. We place the sample on a
planar stage (Marzhauser LStep) and indent vertically using a linear
probe. We use a hemispherical indenter tip with a 6mm diameter printed
in ABS plastic. All indentations are position-controlled relative to
the surface of the sample, which we determine manually by lowering the
probe until we observe contact. While we do not have a force sensor in
our system, we use indentation depth as a proxy for
force. Figure~\ref{fig:instron} shows load vs. indentation depth for
our PDMS samples, determined separately using an Instron testing
machine.

For indentation locations, we use two patterns. The \textit{grid
  indentation pattern} consists of a regular 2D grid of indentation
locations, spaced 2mm apart along each axis. However, the order in
which grid locations were indented was randomized. This is in contrast
with the \textit{random indentation pattern}, where the locations of
the indentations were sampled randomly over the surface of the sample
(not following any pattern).

For each indentation location, we sampled the signal from each pair of
electrodes at multiple indentation depths. Each such measurement
resulted in a tuple of the form $\Phi_i=(x_i,y_i,d_i,r_i^1,..,r_i^6)$,
where $x_i$ and $y_i$ represent the location of the indentation, $d_i$
is the indentation depth, and $r_i^1,..,r_i^6$ (also referred to
collectively as $\bm{r}_i$) represent the change in the six resistance
values we measure between depth $d_i$ and depth 0 (the probe on the
surface of the sample). These tuples are used for data analysis as
described in the next section.

\section{Analysis and Results}

Our main goal is to learn the mapping from all terminal pairs readings
$\bm{r}_i$ to the indentation location $(x,y)$. To train the
predictor, we collected four data sets in regular grid patterns,
totaling 216 indentations. For testing, we collected a dataset
consisting of 60 indentations in a random pattern. All indentations
were performed to a depth of 3mm, or 50\% of the total depth of the
sample. The metric used to quantify the success of this mapping is the
magnitude of the error (in mm) between the predicted indentation
position and ground truth. In our analysis below, we report this error for
individual test points, as well as its mean, median and standard
deviation over the complete testing set.

The baseline that we compare our results against includes a ``Center
Predictor'' and a ``Random Predictor''. The former will always predict
the location of the indentation on the center of our sample, and the
later will predict a completely random location within the sample
surface. The useful area of our sample is 16mm by 10mm; on our test
set, the Center Predictor produces a median error of 5mm, while the
random predictor, if given a large test set, converges on a median
error of above 6mm.

We first attempted Linear Regression as our learning method. The
results were significantly better than the baseline, with a median
error of under 2mm. Still, visual inspection of the magnitude and
direction of the errors revealed a circular bias towards the center
that we attempted to compensate for with a different choice of
learning algorithm. The second regression algorithm we tested was
Ridge Regression with a Laplacian kernel. The Laplacian kernel is a
simple variation of the ubiquitous radial basis kernel, which explains
its ability to remove the non-linear bias noticed in linear regression
results. In this case, we used the first half of the training data for
training the predictor, and the second half to calibrate the ridge
regression tuning factor $\lambda$ and the kernel bandwidth $\sigma$
through grid search.

\begin{figure}[t]
\centering
\begin{tabular}{c}
\includegraphics[width=0.9\linewidth]{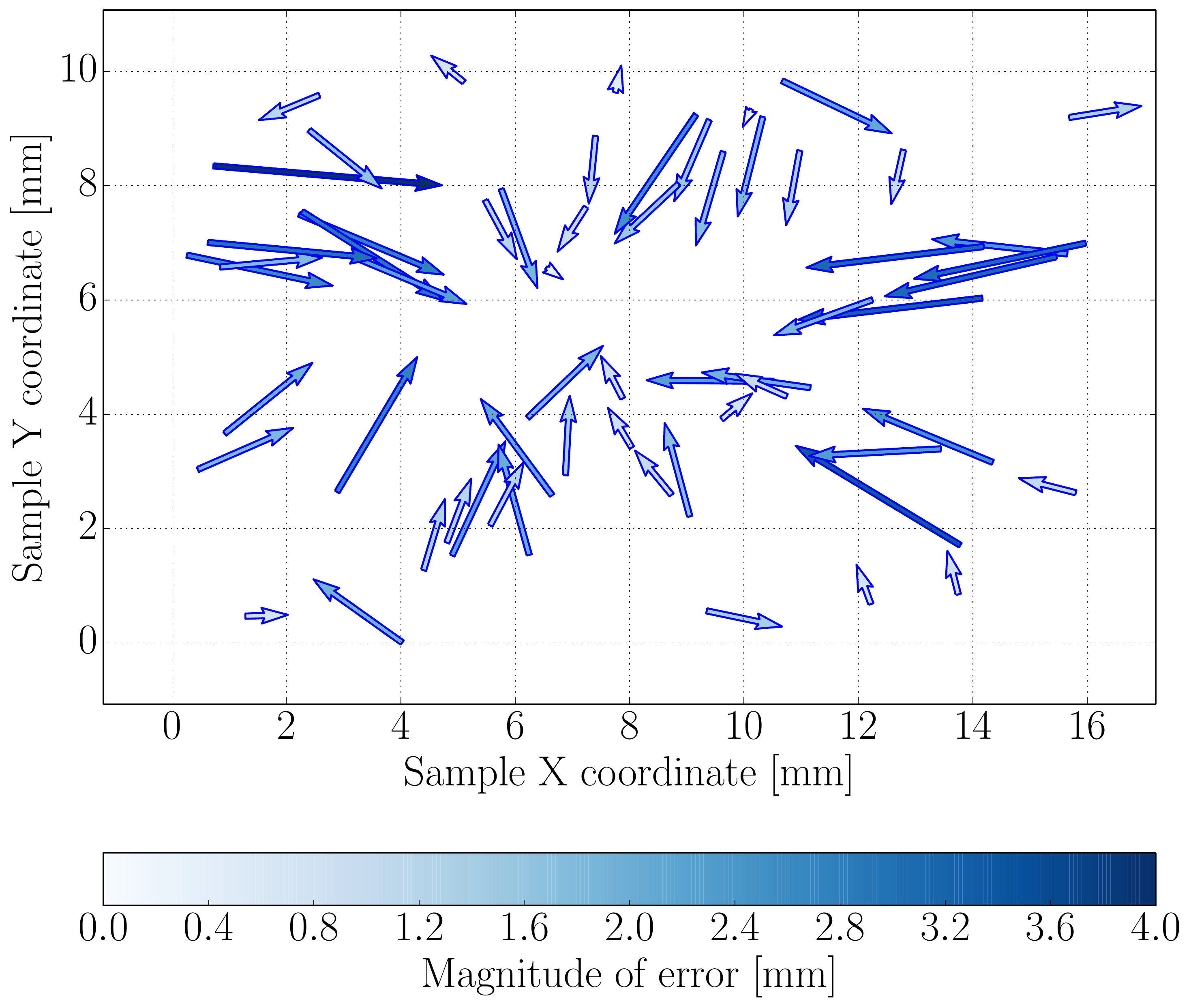}\\
{\footnotesize a) Linear regression}\\[3mm]
\includegraphics[width=0.9\linewidth]{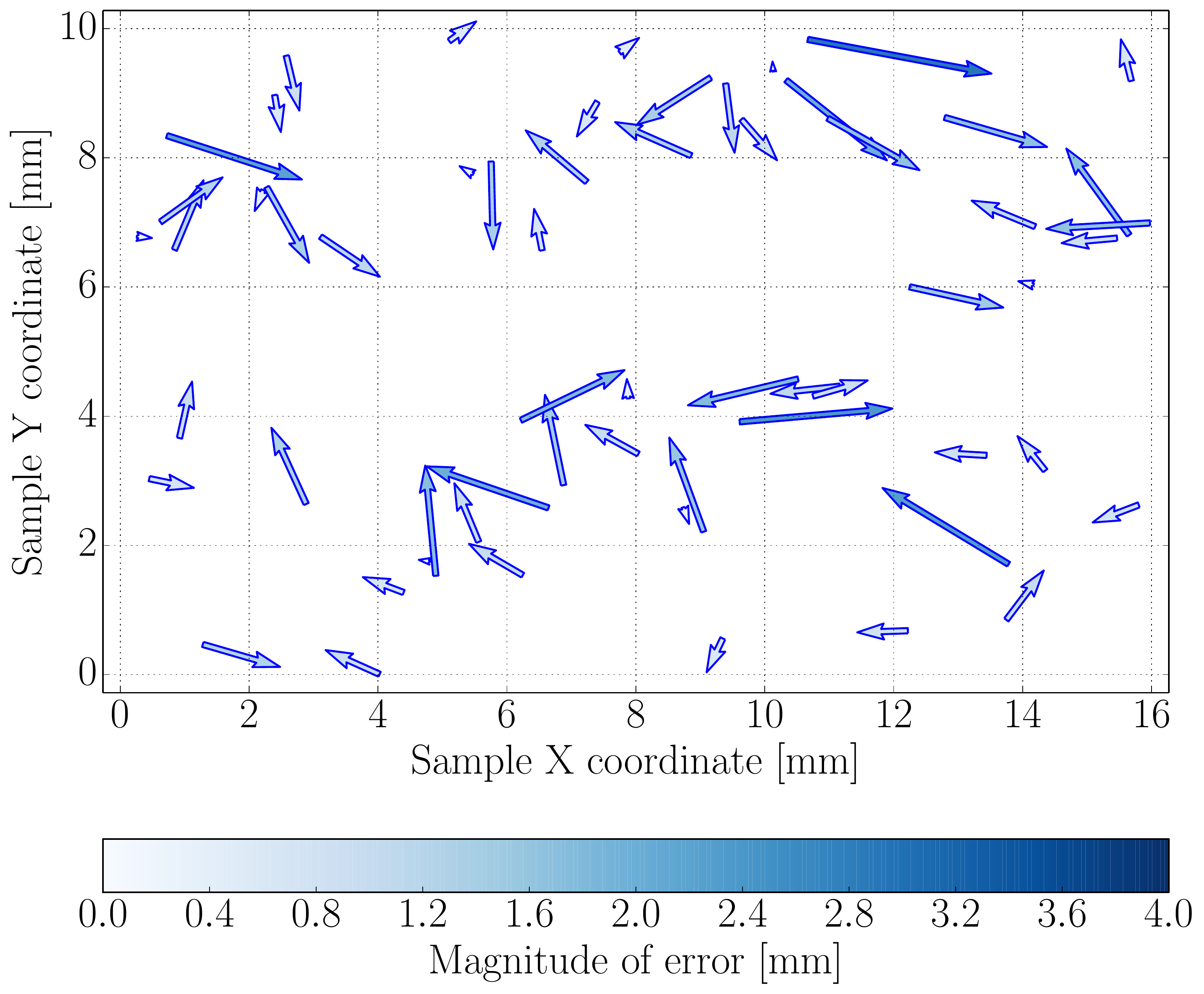}\\
{\footnotesize b) Laplacian ridge regression}
\end{tabular}
\caption{Magnitude and direction of localization errors for Linear and
  Laplacian ridge regressions. Each arrow represents one indentation
  in our test set; the base of the arrow is at the ground truth
  indentation location while the tip is at the predicted location.}
\label{fig:vector}
\vspace{-0mm}
\end{figure}

\begin{figure}[t]
\centering
\begin{tabular}{c}
\includegraphics[width=0.9\linewidth]{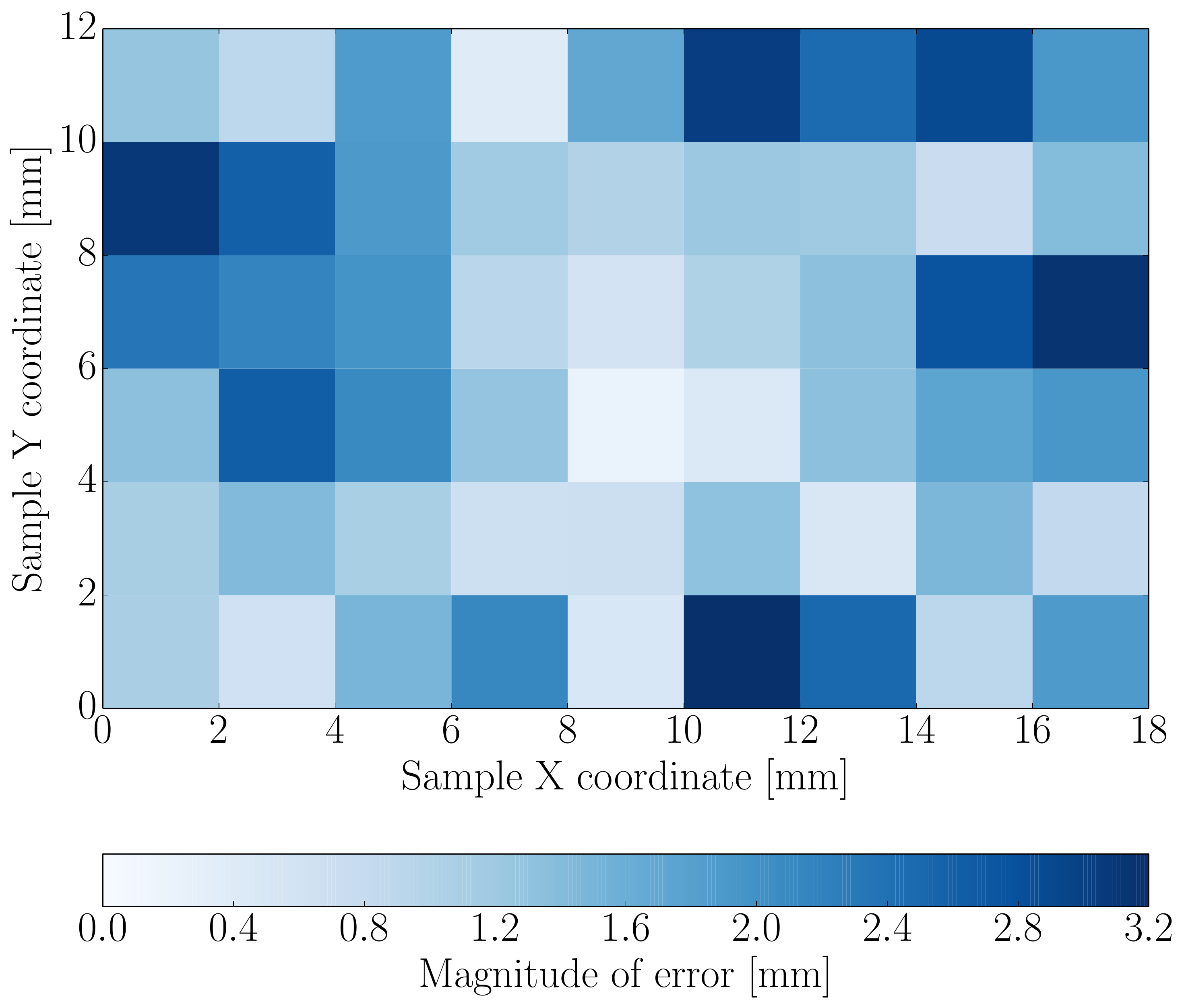}\\
{\footnotesize a) Linear regression}\\[3mm]
\includegraphics[width=0.9\linewidth]{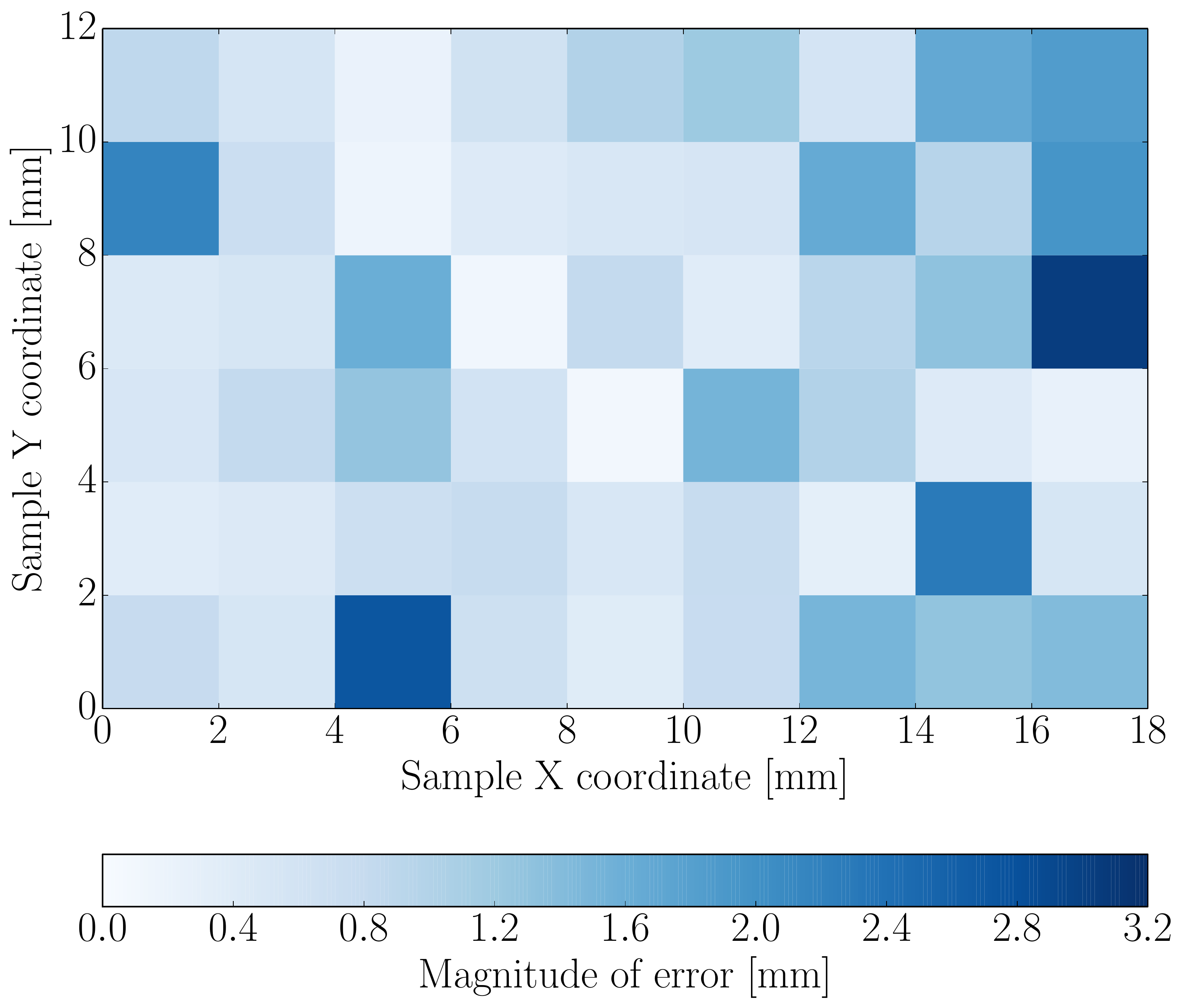}\\
{\footnotesize b) Laplacian ridge regression}
\end{tabular} 
\caption{Heatmap of localization error magnitude based on indentation
  location for Linear and Laplacian ridge regression.}
\label{fig:heat}
\end{figure}

The numerical results using both of our predictors, as well as the two
baseline predictors, are summarized in Table~\ref{table1}. These
results are aggregated over the complete test set consisting of 60
indentations. Linear regression identifies the location of the
indentation within 2mm on average, while Laplacian ridge regression
($\lambda=2.7e^{-2}$, $\sigma=6.15e^{-4}$) further improves this results
achieving sub-millimeter median accuracy.

In addition to the aggregate results, Figure~\ref{fig:vector}
illustrates the magnitude and direction of the localization error for
the entire test set. To characterize localization error uniformly over
the entire sample, we also performed a separate analysis where the
test set also consisted of a regular grid of indentations (in this
case, we used only three such grids for training and the fourth one
for testing). This allows us to plot localization error as a function
of position on the surface of the sample; the results are shown in
Figure~\ref{fig:heat}. Again the predictor using Laplacian ridge
regression achieves high accuracy throughout most of the sample's
area, with larger errors occurring on the edges. We believe that this
pattern can be explained by the fact that an indentation closer to the
center is likely to produce a meaningful signal for more electrode
pairs compared to an indentation at the edge.

\begin{table}[!ht]
\centering
\caption{Prediction accuracy for indentation location}
\label{table1}
\begin{tabular}{lccc}
\hline
\\[-3mm]
\multicolumn{1}{l}{\textbf{Predictor}} & \textbf{Median Err.} & \textbf{Mean Err.} & \textbf{Std. Dev.} \\ \hline
\\[-3mm] \hline
\\[-2mm]
Center predictor                & 5.00 mm      & 5.13 mm    & 2.00 mm        \\ 
Random predictor                & 6.30 mm      & 6.70 mm    & 3.80 mm        \\ 
Linear regression               & 1.75 mm      & 1.75 mm    & 0.83 mm        \\ 
\textbf{Laplacian ridge regression}      & \textbf{0.97 mm}      & \textbf{1.09 mm}    & \textbf{0.59 mm}        \\ 
\end{tabular}
\end{table}

\section{Discussion and Conclusions}

Overall, the results support our hypothesis: that we can achieve high
accuracy spatial resolution for contact determination over a large
sensor area based on a small number of signals collected from
spatially overlapping electrode pairs. The proof of concept sensor,
built as a rectangular shape with an effective sensing area of 10mm by
16mm, discriminates contact location with submillimeter median
accuracy, the equivalent of 160 individual taxels. Even assuming
worst-case accuracy throughout the sensor, we can still locate contact
within 3mm, the equivalent of 15 taxels. This is achieved by measuring
resistance change between 6 electrode pairs, provided by only 4 wires.

Our approach is based on two key ideas. First, the sensor is simply a
continuous volume of piezoresistive material; this allows us to
measure resistance in response to strain between any pair of
electrodes embedded within. On its upper bound, the number of such
pairs is quadratic in the number of wires, and any indentation is
likely to excite a multitude of these signals. Second, we do not
attempt to model analytically the resulting many-to-many mapping
between variables of interest and signals; rather, if such a mapping
exists, we believe it can be learned directly from data.

This proof of concept study is meant to illustrate the feasibility of
this general approach; as such, there are numerous areas of
improvement. Perhaps the most important one will be the ability to
also discriminate contact force, or, in our case, indentation
depth. Additional variables of interest can include planar shear
forces, torsional friction, etc. There are other important aspects
that this initial study does not address, such as repeatability,
hysteresis, lifespan, sensitivity to environmental factors, etc. We
believe that many of these will be determined by the properties of the
underlying transducing modality, which has been extensively studied in
the literature; however, future work will explicitly investigate these
aspects.

Ultimately, the number of variables that can be determined, and the
accuracy that they can be determined with, will depend on the raw data
that can be harvested from the sensor. In this example, we have
demonstrated initial results using 4 electrodes creating 6 unique
pairs, but the number of pairs increases fast with the number of
electrodes (8 electrodes produce 28 pairs, 12 electrodes yield 66
pairs, etc.). Of course, not all electrode pairs will be sensitive to
all indentations, especially if distributed over a large area. Still,
these results lead us to believe that it is possible to capture a rich
description of the contact using this method. Different methods might
also be appropriate for learning the resulting mapping, with new
deep learning approaches as prime candidates. We aim to investigate
these possibilities in future work.

\bibliographystyle{IEEEtran}
\bibliography{bib/tactile,bib/grasping,bib/thesis}

\end{document}